\documentclass{article}
\usepackage{spconf,amsmath,graphicx}
\usepackage{amsfonts}
\usepackage{comment}

\title{Iterative Policy Learning in End-to-End Trainable \\ Task-Oriented Neural Dialog Models}
\name{{\em Bing Liu$^1$, Ian Lane$^1$$^,$$^2$}}
\address{
  $^1$Electrical and Computer Engineering, Carnegie Mellon University \\
  $^2$Language Technologies Institute, Carnegie Mellon University \\
 {\small \tt liubing@cmu.edu, lane@cmu.edu}
}
\begin{document}
%
\maketitle
\begin{abstract}
In this paper, we present a deep reinforcement learning (RL) framework for iterative dialog policy optimization in end-to-end task-oriented dialog systems. Popular approaches in learning dialog policy with RL include letting a dialog agent to learn against a user simulator. Building a reliable user simulator, however, is not trivial, often as difficult as building a good dialog agent. We address this challenge by jointly optimizing the dialog agent and the user simulator with deep RL by simulating dialogs between the two agents. We first bootstrap a basic dialog agent and a basic user simulator by learning directly from dialog corpora with supervised training. We then improve them further by letting the two agents to conduct task-oriented dialogs and iteratively optimizing their policies with deep RL. 
Both the dialog agent and the user simulator are designed with neural network models that can be trained end-to-end.
Our experiment results show that the proposed method leads to promising improvements on task success rate and total task reward comparing to supervised training and single-agent RL training baseline models. 
\end{abstract}
\begin{keywords}
dialog systems, task-oriented, dialog policy, end-to-end, reinforcement learning
\end{keywords}

\section{Introduction}
Task-oriented dialog system is playing an increasingly important role in enabling human-computer interactions via natural spoken language. Different from chatbot type of conversational agents \cite{shang2015neural,serban2016building,li2016persona}, task-oriented dialog systems assist users to complete everyday tasks, which usually involves aggregating information from external resources and planning over multiple dialog turns. Conventional task-oriented dialog systems are designed with complex pipelines \cite{rudnicky1999creating,young2006using,raux2005let,young2013pomdp}, and there are usually separately developed modules for natural language understanding (NLU) \cite{mesnil2015using,chen2016end,Liu2017,liu-lane:2016:SIGDIAL}, dialog state tracking (DST) \cite{henderson2014word,mrkvsic-EtAl:2017:Long}, and dialog management (DM) \cite{gasic2014gaussian,su2016line,su2017sample}. Such pipeline approach inherently make it hard to scale a dialog system to new domains, as each of these modules has to be redesigned separately with domain expertise. Moreover, credit assignment in such pipeline systems can be challenging, as errors made in upper stream modules may propagate and be amplified in downstream components.

Recent efforts have been made in designing end-to-end frameworks for task-oriented dialogs. Wen et al. \cite{wenN2N16} and Liu et al. \cite{Liu+2017} proposed supervised learning (SL) based end-to-end trainable neural network models. Zhao and Eskenazi \cite{zhao2016towards} and Li et al. \cite{li2017end} introduced end-to-end trainable systems using deep reinforcement learning (RL) for dialog policy optimization. Comparing to SL based models, systems trained with RL by exploring the space of possible strategies showed improved model robustness against diverse dialog situations. 

In RL based dialog policy learning, ideally the agent should be deployed to interact with users and get rewards from real user feedback. However, many dialog samples may be required for RL policy shaping, making it impractical to learn from real users directly from the beginning. Therefore, a user simulator \cite{schatzmann2006survey,asri2016sequence,li2016user} is usually used to train the dialog agent to a sufficient starting level before deploying it to the real environment. Quality of such user simulator has a direct impact on the effectiveness of dialog policy learning. Designing a reliable user simulator, however, is not trivial, often as difficult as building a good dialog agent. User simulators used in most of the recent RL based dialog models \cite{zhao2016towards,li2016user,li2017end} are designed with expert knowledge and complex rules. 

To address the challenge of lacking a reliable user simulator for dialog agent policy learning, we propose a method in jointly optimizing the dialog agent policy and the user simulator policy with deep RL. We first bootstrap a basic neural dialog agent and a basic neural user simulator by learning directly from dialog corpora with supervised training. We then improve them further by simulating task-oriented dialogs between the two agents and iteratively optimizing their dialog policies with deep RL. The intuition is that we model task-oriented dialog as a goal fulfilling task, in which we let the dialog agent and the user simulator to positively collaborate to achieve the goal. The user simulator is given a goal to complete, and it is expected to demonstrate coherent but diverse user behavior. The dialog agent attempts to estimate the user's goal and fulfill his request by conducting meaningful conversations. Both the two agents aim to learn to collaborate with each other to complete the task but without exploiting the game. 

Our contribution in this work is two-fold. Firstly, we propose an iterative dialog policy learning method that jointly optimizes the dialog agent and the user simulator in end-to-end trainable neural dialog systems. Secondly, we design a novel neural network based user simulator for task-oriented dialogs that can be trained in a data-driven manner without requiring the design of complex rules.

The remainder of the paper is organized as follows. In section 2, we discuss related work on end-to-end trainable task-oriented dialog systems and RL policy learning methods. In section 3, we describe the proposed framework and learning methods in detail. In Section 4, we discuss the experiment setup and analyze the results. Section 5 gives the conclusions. 

\section{Related Work}
Popular approaches for developing task-oriented dialog systems include treating the problem as a partially observable Markov Decision Process (POMDP) \cite{young2013pomdp}. RL methods \cite{gavsic2013line} can be applied to optimize the dialog policy online with the feedback collected via interacting with users. In order to make the RL policy learning tractable, dialog state and system actions have to be carefully designed.  

Recently, people have proposed neural network based methods for task-oriented dialogs, motivated by their superior performance in modeling chit-chat type of conversations \cite{vinyals2015neural,serban2016building,li2016persona,li2016deep}. Bordes and Weston \cite{bordes2016learning} proposed modeling task-oriented dialogs with a reasoning approach using end-to-end memory networks. Their model skips the belief tracking stage and selects the final system response directly from a list of response candidates. Comparing to this approach, our model explicitly tracks dialog belief state over the sequence of turns, as robust dialog state tracking has been shown \cite{jurvcivcek2012reinforcement} to boost the success rate in task completion. Wen et al. \cite{wenN2N16} proposed an end-to-end trainable neural network model with modularity connected system components. This system is trained in supervised manner, and thus may not be robust enough to handle diverse dialog situations due to the limited varieties in dialog corpus. Our system is trained by a combination of SL and deep RL methods, as it is shown that RL training may effectively improved the system robustness and dialog success rate \cite{henderson2008hybrid,li2017end,williams2017hybrid}. Moreover, other than having separated dialog components as in \cite{wenN2N16}, we use a unified network for belief tracking, knowledge base (KB) operation, and dialog management, to fully explore the knowledge that can be shared among different tasks. 

In many of the recent work on using RL for dialog policy learning \cite{zhao2016towards,williams2016end,li2017end}, hand-designed user simulators are used to interact with the dialog agent. Designing a good performing user simulator is not easy. A too basic user simulator as in \cite{zhao2016towards} may only be able to produce short and simple utterances with limited variety, making the final system lack of robustness against noise in real world user inputs. Advanced user simulators \cite{georgila2005learning,li2016user} may demonstrate coherent user behavior, but they typically require designing complex rules with domain expertise. We address this challenge using a hybrid learning method, where we firstly bootstrapping a basic functioning user simulator with SL on human annotated corpora, and continuously improving it together with the dialog agent during dialog simulations with deep RL. 

Jointly optimizing policies for dialog agent and user simulator with RL has also been studied in literature. Chandramohan et al. \cite{chandramohan2014co} proposed  a co-adaptation framework for dialog systems by jointly optimizing the policies for multiple agents. Georgila et al. \cite{georgila2014single} discussed applying multi-agent RL for policy learning in a resource allocation negotiation scenario. Barlier et al. \cite{barlier2015human} modeled non-cooperative task dialog as a stochastic game and learned jointly the strategies of both agents. Comparing to these previous work, our proposed framework focuses on task-oriented dialogs where the user and the agent positively collaborate to achieve the user's goal. More importantly, we work towards building end-to-end models for task-oriented dialogs that can handle noises and ambiguities in natural language understanding and belief tracking, which is not taken into account in previous work.


\section{Proposed Framework}
In this section, we first provide a high level description of our proposed framework. We then discuss each module component and the training methods in detail. 

In the supervised pre-training stage, we train the dialog agent and the user simulator separately using task-oriented dialog corpora. In the RL training stage, we simulate dialogs between the two agents. The user simulator starts the conversation based on a sampled user goal. The dialog agent attempts to estimate the user's goal and complete the task with the user simulator by conducting multi-turn conversation. At the end of each simulated dialog, a reward is generated based on the level of task completion. This reward is used to further optimize the dialog policies of the two agents with RL.

\subsection{Dialog Agent}
Figure 1 illustrates the design of the dialog agent. The dialog agent is capable of tracking dialog state, issuing API calls to knowledge bases (KB), and producing corresponding system actions and responses by incorporating the query results, which are key skill sets \cite{bordes2016learning} in conducting task-oriented dialogs. State of the dialog agent is maintained in the LSTM \cite{hochreiter1997long} state and being updated after the processing of each turn. At the $k$th turn of a dialog, the dialog agent takes in (1) the previous agent output encoding $o_{k-1}^{A}$, (2) the user input encoding $o_{k-1}^{U}$, (3) the retrieved KB result encoding $o_{k}^{KB}$, and updates its internal state conditioning on the previous agent state $s_{k-1}^{A}$. With the updated agent state $s_{k}^{A}$, the dialog agent emits (1) a system action $a_k^{A}$, (2) an estimation of the belief state, and (3) a pointer to an entity in the retrieved query results. These outputs are then passed to an NLG module to generate the final agent response.
        \begin{figure}[t]
          \centering
          \includegraphics[width=\linewidth]{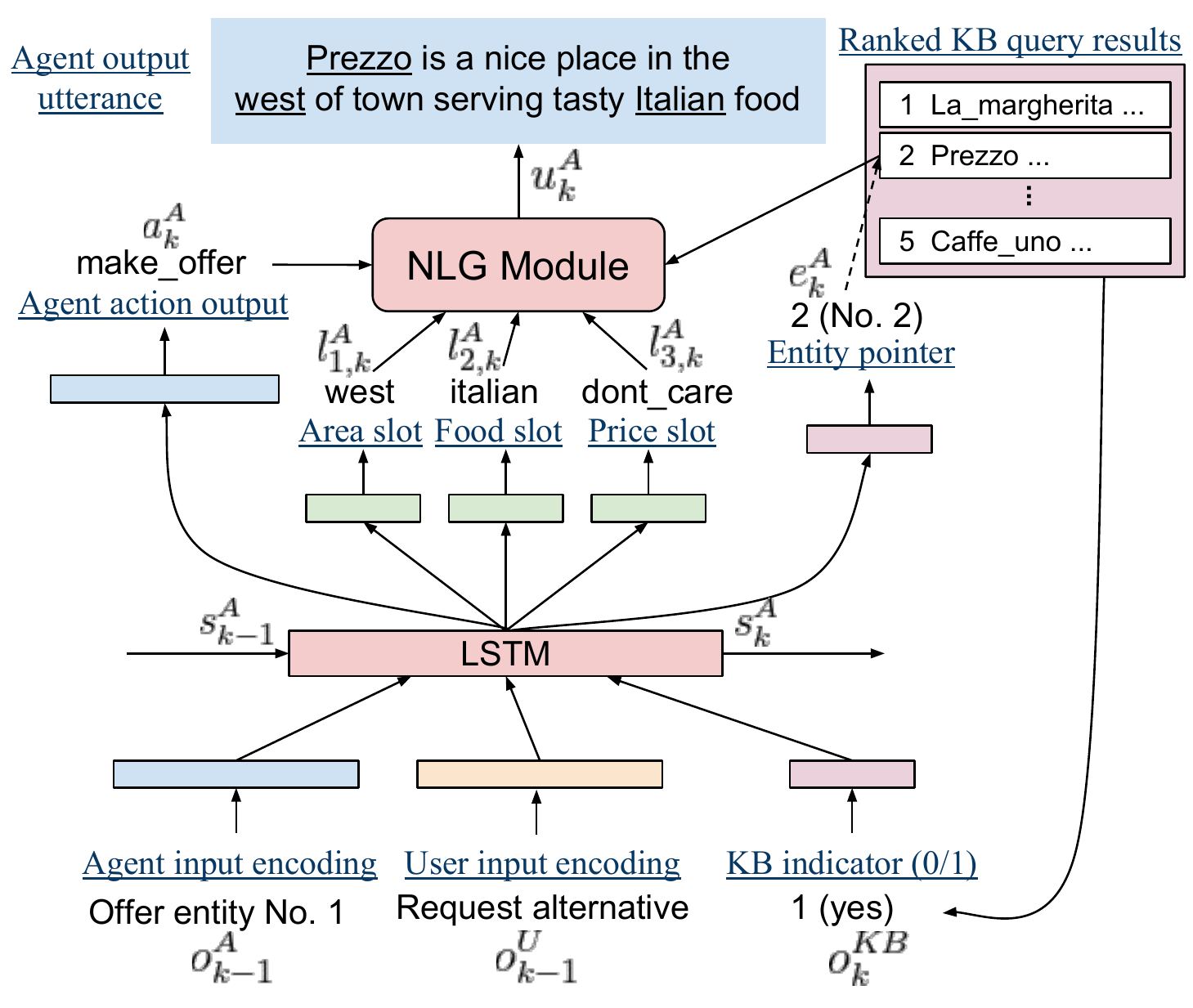}
          \caption{Dialog agent network architecture.}
          \label{fig:dialog_agent}
        \end{figure} 

\textbf{Utterance Encoding} \hspace{3mm} For natural language format inputs at turn $k$, we use bi-directional LSTM to encode the utterance to a continuous vector $o_k$. With $U_k$ representing the utterance at the $k$th turn with $T_k$ words, the utterance vector $o_k$ is produced by concatenating the last forward and backward LSTM states: $o_k =  [\overrightarrow{h^{U_k}_{T_k}}, \overleftarrow{h^{U_k}_{1}}]$. 

\textbf{Action Modeling} \hspace{3mm} We use dialog acts as system actions, which can be seen as shallow representations of the utterance semantics. We treat system action modeling as a multi-class classification problem, where the agent select an appropriate action from a predefined list of system actions based on current dialog state $s_{k}^{A}$: 
    \begin{align}
        & s_{k}^{A} = \operatorname{LSTM_A}(s_{k-1}^{A}, \hspace{1mm} [o_{k-1}^{A}, o_{k-1}^{U}, o_{k}^{KB}]) \\
        & P(a_{k}^{A} \hspace{1mm} | \hspace{1mm} o_{< k}^{A}, \hspace{1mm} o_{< k}^{U}, \hspace{1mm} o_{\le k}^{KB}) = \operatorname{ActDist_A}(s_{k}^{A})
    \end{align}
where $\operatorname{ActDist_A}$ in the agent's network is a multilayer perceptron (MLP) with a single hidden layer and a $\operatorname{softmax}$ activation function over all possible system actions.

\textbf{Belief Tracking} \hspace{3mm} Belief tracker, or dialog state tracker \cite{lee2013structured,henderson2015machine}, continuously tracks the user's goal by accumulating evidence in the conversation. We represent the user's goal using a list of slot values. 
The belief tracker maintains and updates a probability distribution $P(l^{A}_{m,k})$ over candidate values for each slot type $m \in M$ at each turn $k$:
        \begin{align}
            & P(l^{A}_{m,k} \hspace{0.5mm} | \hspace{0.5mm} o_{< k}^{A}, o_{< k}^{U}, o_{\le k}^{KB}) = \operatorname{SlotDist_{A,m}}(s_{k}^{A})
        \end{align}
where $\operatorname{SlotDist}_{m}$ is an MLP with a single hidden layer and a $\operatorname{softmax}$ activation function for slot type $m \in M$.

\textbf{KB Operation} \hspace{3mm} The proposed dialog agent is able to access external information by interfacing with a KB or a database by issuing API calls. Making API call is one of the dialog acts that can be emitted by the agent, conditioning on the state of the conversation. An API call command template is firstly generated with slot type tokens. The final API call command is produced by replacing slot type tokens with the corresponding slot values from the belief tracker outputs.

At the $k$th turn of a dialog, the KB input encoding $o_{k}^{KB}$ is a binary value informing the availability of the entities that match the KB query. Corresponding output is the probability distribution of the entity pointer. 
We treat the KB results as a list of structured entities and let the model to maintain an entity pointer. The agent learns to adjust the entity pointer when user requests for alternative options.

\textbf{Response Generation} \hspace{3mm} We use a template-based NLG module to convert the agent outputs (system action, slot values, and KB entity values) to natural language format.

\subsection{User Simulator}
Figure 2 shows the design of the user simulator. User simulator is given a randomly sampled goal at the beginning of the conversation. Similar to the design of the dialog agent, state of the user simulator is maintained in the state of an LSTM. At the $k$th turn of a dialog, the user simulator takes in (1) the goal encoding $g_{k}^{U}$, (2) the previous user output encoding $o_{k-1}^{U}$, (3) the current turn agent input encoding $o_{k}^{A}$, and updates its internal state conditioning on the previous user state $s_{k-1}^{U}$. On the output side, the user simulator firstly emits a user action $a_{k}^{U}$ based on the updated state $s_{k}^{U}$. Conditioning on this emitted user action and the user dialog state $s_{k}^{U}$, a set of slot values are emitted. The user action and slot values are then passed to an NLG module to generate the final user utterance.
        \begin{figure}[t]
          \centering
          \includegraphics[width=195pt]{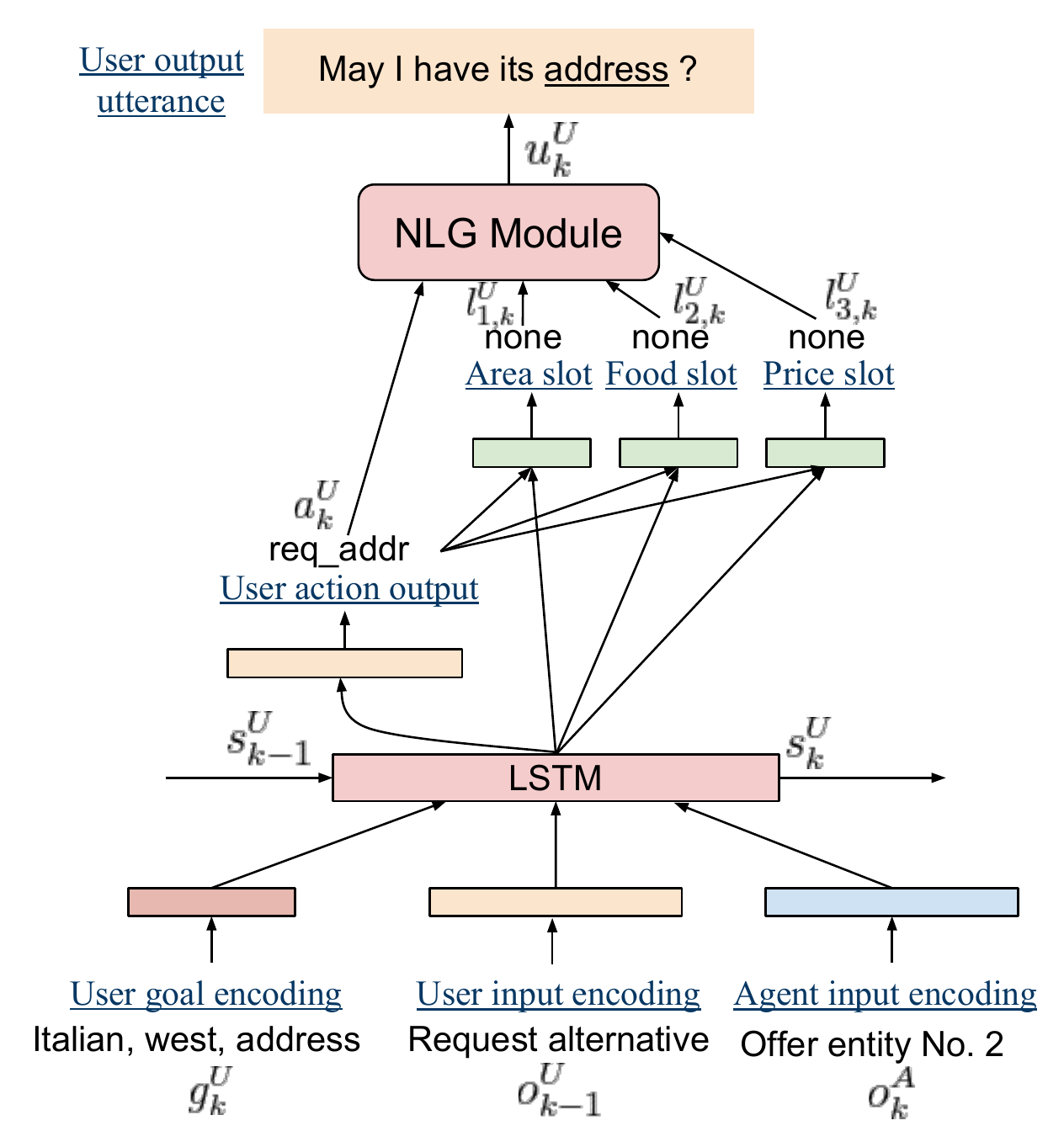}
          \caption{User simulator network architecture.}
          \label{fig:user_simulator}
        \end{figure} 

\textbf{User Goal} \hspace{3mm} We define a user's goal $g^{U}$ using a list of informable and requestable slots \cite{williams2016dialog}. Informable slots are the slots that users can provide a value for to describe their goal (e.g. slots for food type, area, etc.). 
Requestable slots are the slots that users want to request the value for, such as requesting the \textit{address} of a restaurant. We treat informable slots as discrete type of inputs that can take multiple values, and treat requestable slots as inputs that take binary values (i.e. a slot is either requested or not). In this work, once the a goal is sampled at the beginning of the conversation, we fix the user goal and do not change it during the conversation.

\textbf{Action Selection} \hspace{3mm} Similar to the action modeling in dialog agent, we treat user action modeling as a multi-class classification problem conditioning on the dialog context encoded in the dialog-level LSTM state $s_{k}^{U}$ on the user simulator side: 
    \begin{align}
        & s_{k}^{U} = \operatorname{LSTM_U}(s_{k-1}^{U}, \hspace{1mm} [g_{k}^{U}, \hspace{1mm} o_{k-1}^{U}, \hspace{1mm} o_{k}^{A}]) \\
        & P(a_{k}^{U} \hspace{1mm} | \hspace{1mm} g_{\le k}^{U},\hspace{1mm} o_{< k}^{U}, \hspace{1mm} o_{\le k}^{A}) = \operatorname{ActDist_U}(s_{k}^{U})
    \end{align}
Once user action is generated at turn $k$, it is used together with the current user dialog state $s_{k}^{U}$ to generate value for each informable slot:
        \begin{align}
            & P(l^{U}_{m,k} \hspace{0.5mm} | \hspace{0.5mm} g_{\le k}^{U},o_{< k}^{U}, o_{\le k}^{A}) = \operatorname{SlotDist_{U,m}}(s_{k}^{U}, a_{k}^{U})
        \end{align}
Similar to the design of the dialog agent, $\operatorname{ActDist_U}$ and $\operatorname{SlotDist_{U,m}}$ are MLPs with a single hidden layer and use $\operatorname{softmax}$ activation over their corresponding outputs.

\textbf{Utterance Generation} \hspace{3mm} We use a template-based NLG module to convert the user simulator's outputs (action and slot values) to the natural language surface form. 

\subsection{Deep RL Policy Optimization}
        \begin{figure*}[t]
          \centering
          \includegraphics[width=400pt]{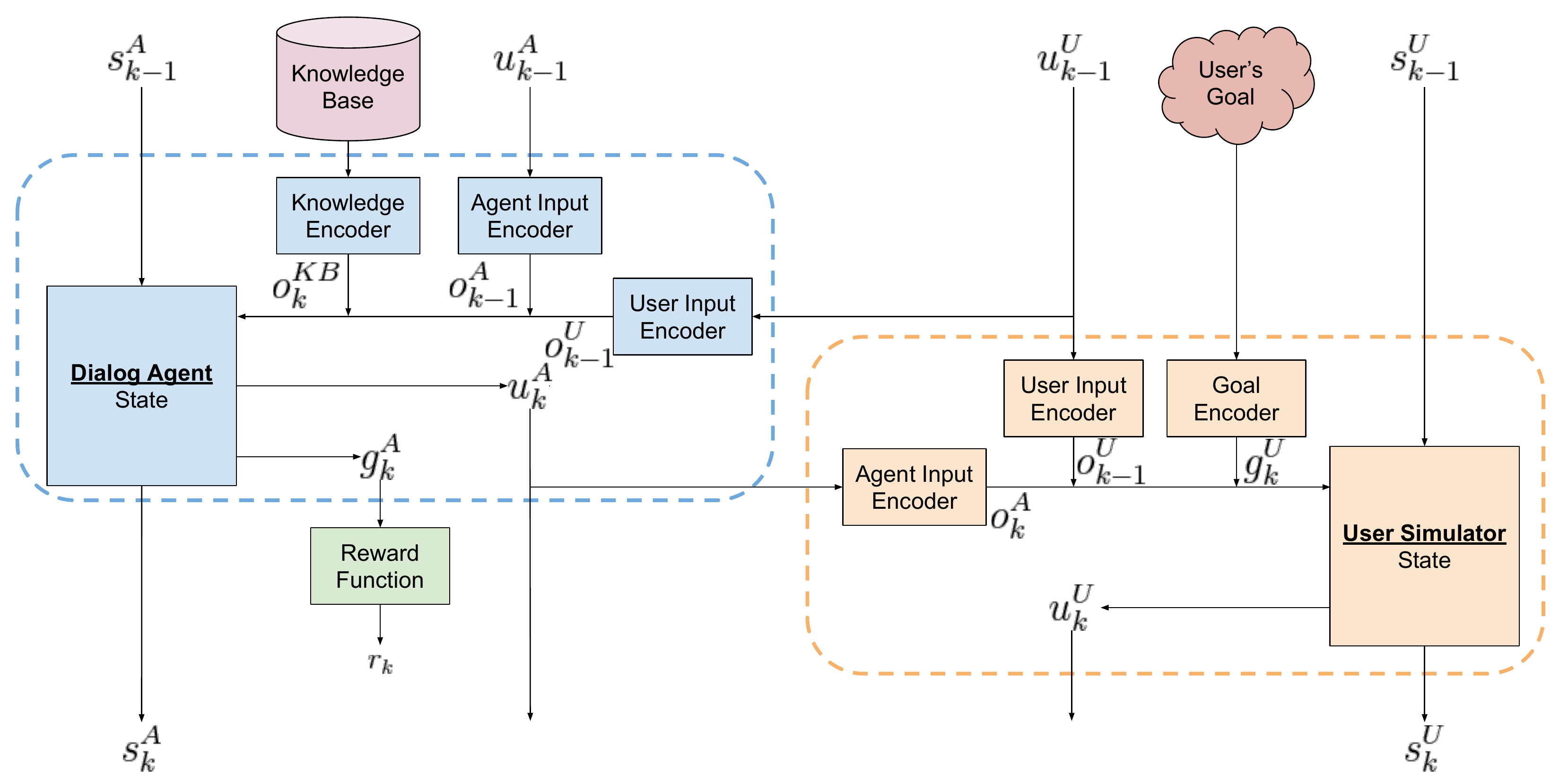}
          \caption{System architecture for joint dialog agent and user simulator policy optimization with deep RL}
          \label{fig:RL_dialog_learning}
        \end{figure*} 
RL policy optimization is performed on top of the supervised pre-trained networks. The system architecture is shown in Figure \ref{fig:RL_dialog_learning}. We defines the state, action, and reward in our RL training setting and present the training details.

\textbf{State} \hspace{3mm} 
For RL policy learning, states of the dialog agent and the user simulator at the $k$th turn are the dialog-level LSTM state $s_{k}^{A}$ and $s_{k}^{U}$ respectively. Both $s_{k}^{A}$ and $s_{k}^{U}$ captures the dialog history up till current turn. The user state $s_{k}^{U}$ also encodes the user's goal.

\textbf{Action} \hspace{3mm} 
Actions of the dialog agent and user simulator are the system action outputs $a_{k}^{A}$ and $a_{k}^{U}$. An action is sampled by the agent based on a stochastic representation of the policy, which produces a probability distribution over actions given a dialog state. Action space is finite and discrete for both the dialog agent and the user simulator. 

\textbf{Reward} \hspace{3mm}  
Reward is calculated based on the level of task completion. A turn level reward $r_k$ is applied based on the progress that the agent and user made in completing the predefined task over the past turn. At the end of each turn, a score $score_k$ is calculated indicating to what extend the agent has fulfilled the user's request so far. The turn level reward $r_k$ is then calculated by the difference of the scores received in two consecutive turns: 
        \begin{align}
            score_k & = \mathcal{D}(g_k^{U}, g_k^{A}) \\
            r_k & = score_k - score_{k-1}
        \end{align}
where $\mathcal{D}$ is the scoring function. $g_k^{U}$ and $g_k^{A}$ are the true user's goal and the agent's estimation of the user's goal, both are represented by slot-value pairs. Alternatively, the turn level reward $r_k$ can be obtained by using the discounted reward received at the end of the dialog (positive reward for task success, and negative reward for task failure). 

\textbf{Policy Gradient RL} \hspace{3mm} 
For policy optimization with RL, policy gradient method is preferred over Q-learning in our system as the policy network parameters can be initialized with the $\operatorname{ActDist}$ parameters learnied during supervised pre-training stage. With REINFORCE \cite{williams1992simple}, the objective function can be written as $J_k(\theta_a, \theta_u) = \mathbb E_{\theta_a, \theta_u}\left[ R_k  \right] = \mathbb E_{\theta_a, \theta_u}\left[ \sum_{t=0}^{K-k} \gamma ^{t}r_{k+t}  \right]$, with $\gamma \in [0,1]$ being the discount factor. We optimize parameter sets $\theta _a$ and $\theta _u$ for the dialog agent and the user simulator to maximize $J_k(\theta_a, \theta_u)$. For the dialog agent, with likelihood ratio gradient estimator, gradient of $J_k(\theta_a, \theta_u)$ can be derived as:
        \begin{align}
            &\nabla  _{\theta _a} J_k(\theta_a, \theta_u) = \nabla _{\theta _a} \mathbb E_{\theta_a, \theta_u}\left[ R_k \right]&\\
            &= \nabla _{\theta _a} \sum_{a_{k}^{a}, a_{k}^{u}}  \pi _{\theta _{a}}(a_{k}^{a} | s_{k}^{a}) \pi _{\theta _{u}}(a_{k}^{u} | s_{k}^{u}) R_{k, a_{k}^{a}, a_{k}^{u}}&\\
            &= \mathbb E_{\theta_a, \theta_u}\left[ \nabla _{\theta _a} \log \pi _{\theta _{a}}(a_{k}^{a} | s_{k}^{a}) R_{k} \right]&
        \end{align} 
This last expression above gives us an unbiased gradient estimator. We sample agent action and user action at each dialog turn and compute the policy gradient. Similarly, gradient on the user simulator side can be derived as:
        \begin{align}
            &\nabla  _{\theta _u} J_k(\theta_a, \theta_u) = \nabla _{\theta _u} \mathbb E_{\theta_a, \theta_u}\left[ R_k \right]&\\
            &= \mathbb E_{\theta_a, \theta_u}\left[ \nabla _{\theta _u} \log \pi _{\theta _{u}}(a_{k}^{u} | s_{k}^{u}) R_{k} \right]&
        \end{align} 
A potential drawback of using REINFORCE is that the policy gradient might have high variance, since the agent may take many steps over the course of a dialog episode. We also explore using Advantage Actor-Critic (A2C) in our study, in which we approximate a state-value function using a feed forward neural network.


During model training, we use softmax policy for both the dialog agent and the user simulator to encourage exploration. Softmax policy samples action from the action probability distribution calculated by the $\operatorname{softmax}$ in the system action output. During evaluation, we apply greedy policy to the dialog agent, and still apply softmax policy to the user simulator. This is to increase randomness and diversity in the user simulator behavior, which is closer to the realistic dialog system evaluation settings with human users. This also prevents the two agents from fully cooperating with each other and exploiting the game.

\section{Experiments}
\subsection{Dataset}
We prepare the data in our study based the corpus from the second Dialog State Tracking Challenge (DSTC2) \cite{henderson2014second}. We converte this corpus to our required format by adding API call commands and the corresponding KB query results. The dialogs simulation is based on real KB search results, which makes the dialog agent evaluation closer to real cases. 
Different from DSTC2, agent and user actions in our system are generated by concatenating the act and slot names in the original dialog act output (e.g. ``$confirm(food=italian)$'' maps to ``$confirm\_food$''). Slot values are captured in the belief tracking outputs. Table 1 shows the statistics of the dataset used in our experiments.
        \begin{table}[ht]
          \label{tab:table_1}
          \centering
          \begin{tabular}{l r}
            \hline
            \# of train/dev/test dialogs           & 1612 / 506 / 1117     \\
            \# of turns per dialog in average    & 7.9 \\ 
            \# of dialog agent actions    & 52 \\ 
            \# of user simulator actions    & 100 \\ 
            \# of area / food / pricerange options       & 5 / 91 / 3 \\
            \hline
          \end{tabular}     
        \caption{Statistics of the dataset}
        \end{table}
\subsection{Training Procedure}
In supervised pre-training, the dialog agent and the user simulator are trained separately against dialog corpus. We use the same set of neural network model configurations for both agents. Hidden layer sizes of the dialog-level LSTM for dialog modeling and utterance-level LSTM for utterance encoding are both set as 150. We perform mini-batch training using Adam optimization method \cite{kingma2014adam}. Initial learning rate is set as 1e-3. Dropout \cite{srivastava2014dropout} ($p=0.5$) is applied during model training to prevent to model from over-fitting.

In deep RL training stage, the policy network parameters are initialized with $\operatorname{ActDist}$ parameters from the SL training. State-value function network parameters in A2C are initialized randomly. To ameliorate the non-stationarity problem when jointly training the two agents, we update the two agents iteratively during RL training. We take 100 episodes as a RL training cycle, in which we fix one agent and only update the other, and switch the training agent in the next cycle until convergence. In dialog simulation, we end the dialog if the dialog length exceeds the maximum turn size (20 in our experiment) or the user simulator emits the end of dialog action. 

\subsection{Results and Analysis}
We evaluate the system on task success rate, average task reward, and average dialog length on simulated dialogs. A dialog is considered successful if the agent's belief tracking outputs match the informable user goal slot values completely, and all user requested slots are fulfilled. Note that the results on task success rate in this work should not be directly compared to the numbers in \cite{wenN2N16,wen2016conditional}, as both the dialog agent and the user simulator in our study are end-to-end models that take noisy natural language utterance as input and directly generate the final dialog act output. Moreover, instead of using greedy policy on user simulator, we sample user actions based on the action probability distribution from the user policy network to encourage diversity and variety in user behavior.

Table 2 shows the evaluation results. The baseline model uses the SL trained agents. REINFORCE-agent and A2C-agent apply RL training on the dialog agent only, without updating the user simulator. REINFORCE-joint and A2C-joint apply RL on both the dialog agent and user simulator over the SL pre-trained models. Figure 4, 5, and 6 show the learning curves of these five models during RL training on dialog success rate, average reward, and average success dialog length. 

        \begin{table}[h]
          \label{tab:table_2}
          \centering
          \begin{tabular}{l c c c}
            \hline
                            & Success          & Avg    & Avg Success\\
                            & Rate (\%)          & Reward   & Turn Size\\
            \hline
            SL Baseline        & 35.3                    & 2.02  & 6.46\\
            REINFORCE-agent         & 50.2                   & 2.96  & 6.47 \\
            REINFORCE-joint         & 61.1                    & \textbf{3.30}  & 5.12\\
            A2C-agent        & 50.6                    & 3.11  & 6.03\\
            A2C-joint        & \textbf{64.7}          & 3.23  & 6.71 \\
            \hline
          \end{tabular}     
        \caption{Evaluation results on the converted DSTC2 dataset.}
        \end{table}        

        \begin{figure}[h]
          \centering
          \includegraphics[width=\linewidth]{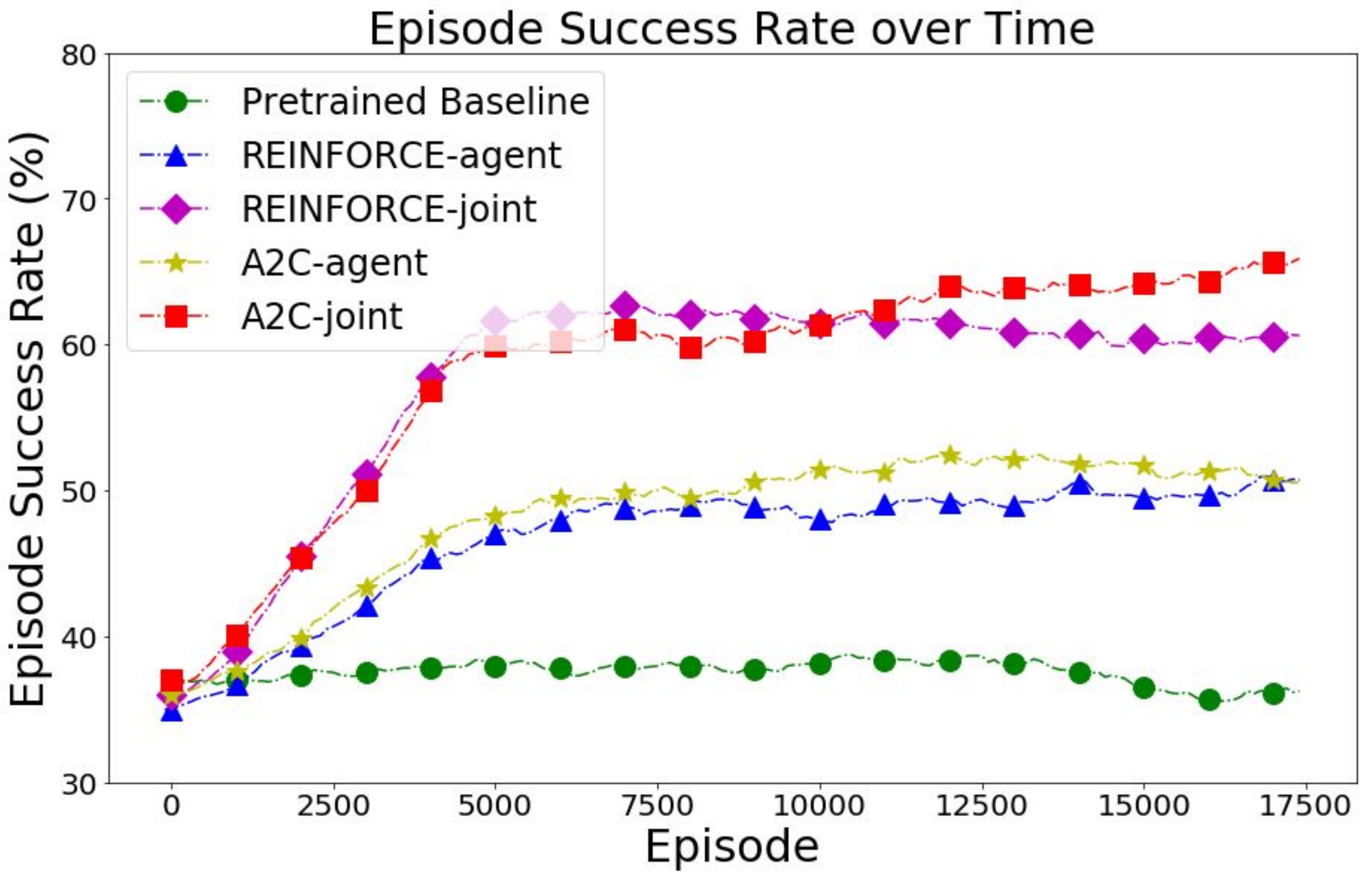}
          \caption{Learning curve for task success rate.}
          \label{fig:success_rate}
        \end{figure}
        
\textbf{Success Rate} \hspace{3mm} 
As shown in Table 2, the SL model achieves the lowest task success rate. Model trained with SL on dialog corpus has limited capabilities in capturing the change in state, and thus may not be able to generalize well to unseen dialog situations during simulation. RL efficiently improves the dialog task success rate, as it enables the dialog agent to explore strategies that are not in the training corpus. The agent-update-only models using REINFORCE and A2C achieve similar results, outperforming the baseline model by 14.9\% and 15.3\% respectively. The jointly optimized models improved the performance further over the agent-update-only models. Model using A2C for joint policy optimization achieves the best task success rate. 

        \begin{figure}[h]
          \centering
          \includegraphics[width=\linewidth]{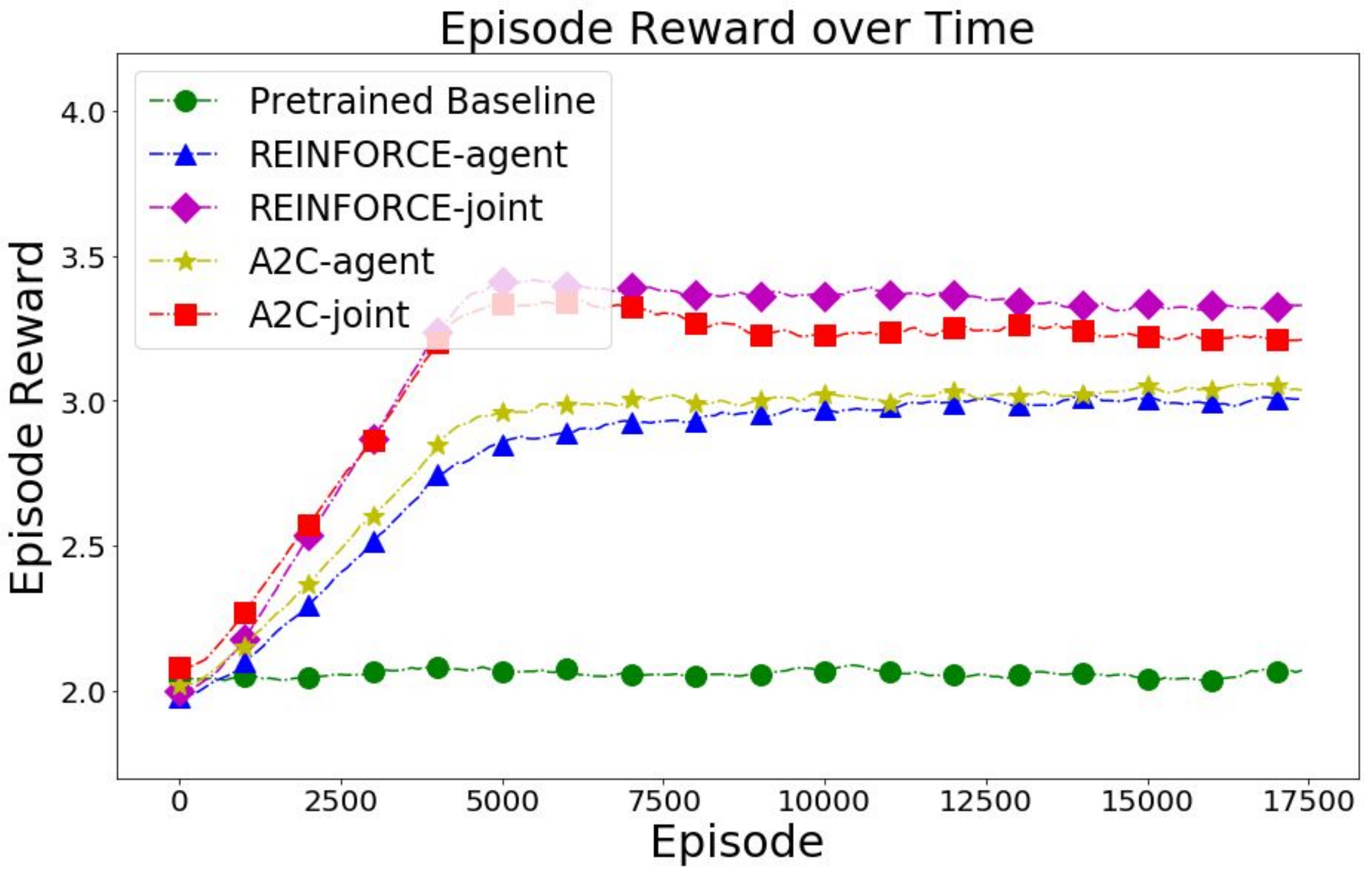}
          \caption{Learning curve for average reward}
          \label{fig:reward}
        \end{figure}
\textbf{Average Reward} \hspace{3mm} 
RL curves on average dialog reward show similar trends as above. One difference is that the joint training model using REINFORCE achieves the highest average reward, outperforming that using A2C by a small margin. This is likely due to the better performance of our REINFORCE models in earning reward in the failed dialogs. We find that our user simulator trained with A2C tends to have sharper action distribution from the softmax policy, making it easier to get stuck when it falls into an unfavorable state. We are interested in exploring fine-grained control strategies in joint RL policy optimization framework in our future work.

        \begin{figure}[h]
          \centering
          \includegraphics[width=\linewidth]{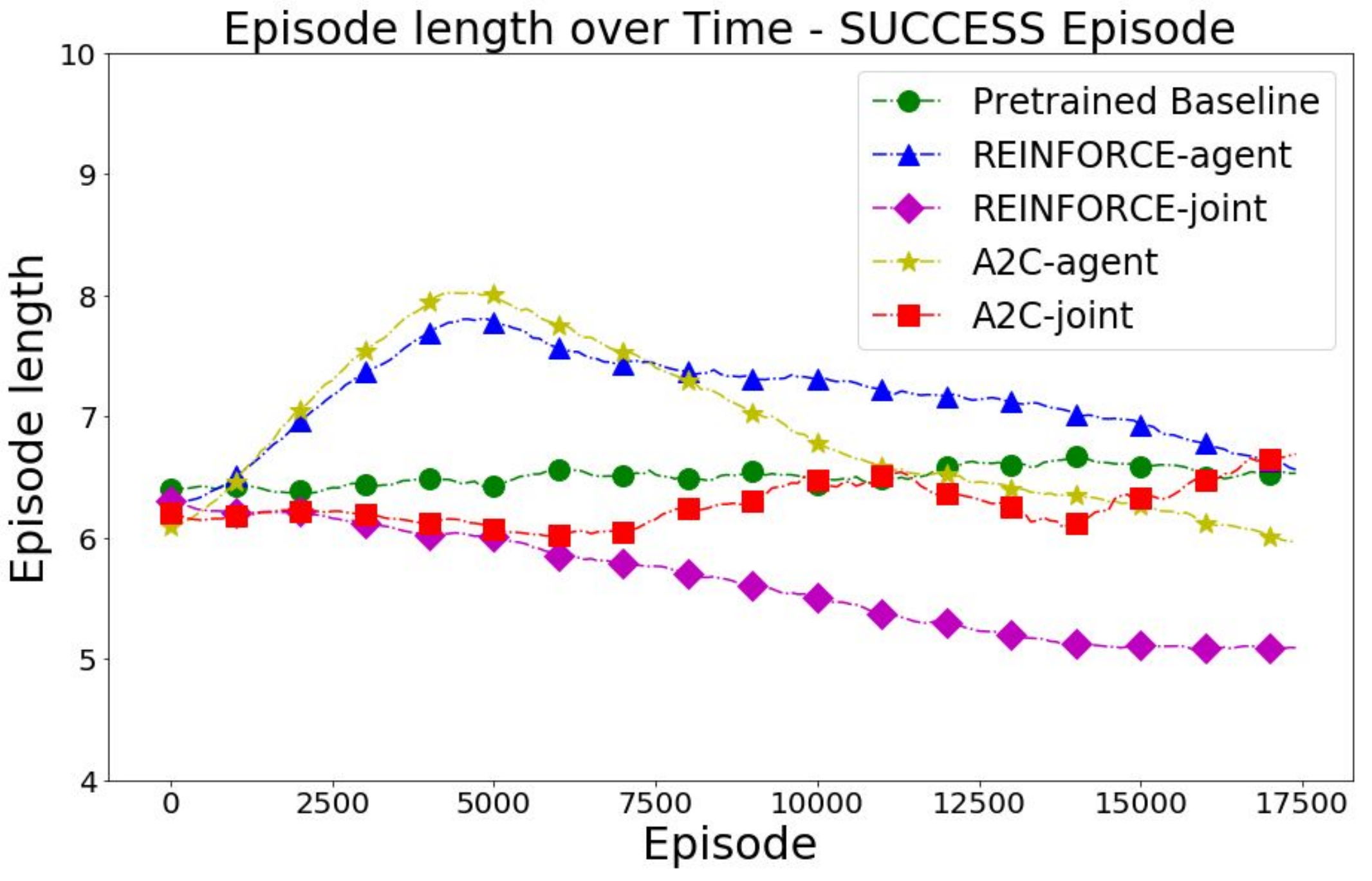}
          \caption{Learning curve for average turn size}
          \label{fig:dialog_length}
        \end{figure}
\textbf{Average Success Turn Size} \hspace{3mm} 
The average turn size of success dialogs tends to decrease along the episode of RL policy learning. This is in line with our expectation as both the dialog agent and the user simulator improve their policies for more efficient and coherent strategies with the RL training. 

\section{Conclusions}
In this work, we propose a reinforcement learning framework for dialog policy optimization in end-to-end task-oriented dialog systems. The proposed method addresses the challenge of lacking a reliable user simulator for policy learning in task-oriented dialog systems. We present an iterative policy learning method that jointly optimizes the dialog agent and the user simulator with deep RL by simulating dialogs between the two agents. Both the dialog agent and the user simulator are designed with neural network models that can be trained end-to-end. 
Experiment results show that our proposed method leads to promising improvements on task success rate and task reward over the supervised training and single-agent RL training baseline models.

\bibliographystyle{IEEEbib}
\bibliography{strings}

\begin{thebibliography}{10}

\bibitem{shang2015neural}
Lifeng Shang, Zhengdong Lu, and Hang Li,
\newblock ``Neural responding machine for short-text conversation,''
\newblock in {\em ACL-IJCNLP}, 2015.

\bibitem{serban2016building}
Iulian~V Serban, Alessandro Sordoni, Yoshua Bengio, Aaron Courville, and Joelle
  Pineau,
\newblock ``Building end-to-end dialogue systems using generative hierarchical
  neural network models,''
\newblock in {\em Proceedings of the 30th AAAI Conference on Artificial
  Intelligence (AAAI-16)}, 2016.

\bibitem{li2016persona}
Jiwei Li, Michel Galley, Chris Brockett, Georgios~P Spithourakis, Jianfeng Gao,
  and Bill Dolan,
\newblock ``A persona-based neural conversation model,''
\newblock in {\em ACL}, 2016.

\bibitem{rudnicky1999creating}
Alexander~I Rudnicky, Eric~H Thayer, Paul~C Constantinides, Chris Tchou,
  R~Shern, Kevin~A Lenzo, Wei Xu, and Alice Oh,
\newblock ``Creating natural dialogs in the carnegie mellon communicator
  system.,''
\newblock in {\em Eurospeech}, 1999.

\bibitem{young2006using}
Steve Young,
\newblock ``Using pomdps for dialog management,''
\newblock in {\em Spoken Language Technology Workshop, 2006. IEEE}. IEEE, 2006,
  pp. 8--13.

\bibitem{raux2005let}
Antoine Raux, Brian Langner, Dan Bohus, Alan~W Black, and Maxine Eskenazi,
\newblock ``Let’s go public! taking a spoken dialog system to the real
  world,''
\newblock in {\em in Proc. of Interspeech 2005}. Citeseer, 2005.

\bibitem{young2013pomdp}
Steve Young, Milica Ga{\v{s}}i{\'c}, Blaise Thomson, and Jason~D Williams,
\newblock ``Pomdp-based statistical spoken dialog systems: A review,''
\newblock {\em Proceedings of the IEEE}, vol. 101, no. 5, pp. 1160--1179, 2013.

\bibitem{mesnil2015using}
Gr{\'e}goire Mesnil, Yann Dauphin, Kaisheng Yao, Yoshua Bengio, Li~Deng, Dilek
  Hakkani-Tur, Xiaodong He, Larry Heck, Gokhan Tur, Dong Yu, et~al.,
\newblock ``Using recurrent neural networks for slot filling in spoken language
  understanding,''
\newblock {\em IEEE/ACM Transactions on Audio, Speech and Language Processing
  (TASLP)}, vol. 23, no. 3, pp. 530--539, 2015.

\bibitem{chen2016end}
Yun-Nung Chen, Dilek Hakkani-T{\"u}r, G{\"o}khan T{\"u}r, Jianfeng Gao, and
  Li~Deng,
\newblock ``End-to-end memory networks with knowledge carryover for multi-turn
  spoken language understanding.,''
\newblock in {\em INTERSPEECH}, 2016, pp. 3245--3249.

\bibitem{Liu2017}
Bing Liu and Ian Lane,
\newblock ``An end-to-end trainable neural network model with belief tracking
  for task-oriented dialog,''
\newblock in {\em Interspeech}, 2017, pp. 2506--2510.

\bibitem{liu-lane:2016:SIGDIAL}
Bing Liu and Ian Lane,
\newblock ``Joint online spoken language understanding and language modeling
  with recurrent neural networks,''
\newblock in {\em SIGDIAL}, 2016, pp. 22--30.

\bibitem{henderson2014word}
Matthew Henderson, Blaise Thomson, and Steve Young,
\newblock ``Word-based dialog state tracking with recurrent neural networks,''
\newblock in {\em SIGDIAL}, 2014, pp. 292--299.

\bibitem{mrkvsic-EtAl:2017:Long}
Nikola Mrk\v{s}i\'{c}, Diarmuid \'{O}~S\'{e}aghdha, Tsung-Hsien Wen, Blaise
  Thomson, and Steve Young,
\newblock ``Neural belief tracker: Data-driven dialogue state tracking,''
\newblock in {\em ACL}, 2017, pp. 1777--1788.

\bibitem{gasic2014gaussian}
Milica Gasic and Steve Young,
\newblock ``Gaussian processes for pomdp-based dialogue manager optimization,''
\newblock {\em IEEE/ACM Transactions on Audio, Speech, and Language
  Processing}, vol. 22, no. 1, pp. 28--40, 2014.

\bibitem{su2016line}
Pei-Hao Su, Milica Gasic, Nikola Mrksic, Lina Rojas-Barahona, Stefan Ultes,
  David Vandyke, Tsung-Hsien Wen, and Steve Young,
\newblock ``On-line active reward learning for policy optimisation in spoken
  dialogue systems,''
\newblock in {\em ACL}, 2016.

\bibitem{su2017sample}
Pei-Hao Su, Pawel Budzianowski, Stefan Ultes, Milica Gasic, and Steve Young,
\newblock ``Sample-efficient actor-critic reinforcement learning with
  supervised data for dialogue management,''
\newblock in {\em SIGDIAL}, 2017.

\bibitem{wenN2N16}
Tsung-Hsien Wen, David Vandyke, Nikola Mrk{\v{s}}i\'c, Milica Ga{\v{s}}i\'c,
  Lina M.~Rojas-Barahona, Pei-Hao Su, Stefan Ultes, and Steve Young,
\newblock ``A network-based end-to-end trainable task-oriented dialogue
  system,''
\newblock in {\em EACL}, 2017.

\bibitem{Liu+2017}
Bing Liu and Ian Lane,
\newblock ``An end-to-end trainable neural network model with belief tracking
  for task-oriented dialog,''
\newblock in {\em Interspeech}, 2017.

\bibitem{zhao2016towards}
Tiancheng Zhao and Maxine Eskenazi,
\newblock ``Towards end-to-end learning for dialog state tracking and
  management using deep reinforcement learning,''
\newblock in {\em SIGDIAL}, 2016.

\bibitem{li2017end}
Xuijun Li, Yun-Nung Chen, Lihong Li, and Jianfeng Gao,
\newblock ``End-to-end task-completion neural dialogue systems,''
\newblock {\em arXiv preprint arXiv:1703.01008}, 2017.

\bibitem{schatzmann2006survey}
Jost Schatzmann, Karl Weilhammer, Matt Stuttle, and Steve Young,
\newblock ``A survey of statistical user simulation techniques for
  reinforcement-learning of dialogue management strategies,''
\newblock {\em The knowledge engineering review}, vol. 21, no. 2, pp. 97--126,
  2006.

\bibitem{asri2016sequence}
Layla~El Asri, Jing He, and Kaheer Suleman,
\newblock ``A sequence-to-sequence model for user simulation in spoken dialogue
  systems,''
\newblock in {\em Interspeech}, 2016.

\bibitem{li2016user}
Xiujun Li, Zachary~C Lipton, Bhuwan Dhingra, Lihong Li, Jianfeng Gao, and
  Yun-Nung Chen,
\newblock ``A user simulator for task-completion dialogues,''
\newblock {\em arXiv preprint arXiv:1612.05688}, 2016.

\bibitem{gavsic2013line}
M~Ga{\v{s}}i{\'c}, Catherine Breslin, Matthew Henderson, Dongho Kim, Martin
  Szummer, Blaise Thomson, Pirros Tsiakoulis, and Steve Young,
\newblock ``On-line policy optimisation of bayesian spoken dialogue systems via
  human interaction,''
\newblock in {\em ICASSP}. IEEE, 2013, pp. 8367--8371.

\bibitem{vinyals2015neural}
Oriol Vinyals and Quoc Le,
\newblock ``A neural conversational model,''
\newblock in {\em ICML}, 2015.

\bibitem{li2016deep}
Jiwei Li, Will Monroe, Alan Ritter, Michel Galley, Jianfeng Gao, and Dan
  Jurafsky,
\newblock ``Deep reinforcement learning for dialogue generation,''
\newblock in {\em EMNLP}, 2016.

\bibitem{bordes2016learning}
Antoine Bordes and Jason Weston,
\newblock ``Learning end-to-end goal-oriented dialog,''
\newblock in {\em ICLR}, 2017.

\bibitem{jurvcivcek2012reinforcement}
Filip Jur{\v{c}}{\'\i}{\v{c}}ek, Blaise Thomson, and Steve Young,
\newblock ``Reinforcement learning for parameter estimation in statistical
  spoken dialogue systems,''
\newblock {\em Computer Speech \& Language}, vol. 26, no. 3, pp. 168--192,
  2012.

\bibitem{henderson2008hybrid}
James Henderson, Oliver Lemon, and Kallirroi Georgila,
\newblock ``Hybrid reinforcement/supervised learning of dialogue policies from
  fixed data sets,''
\newblock {\em Computational Linguistics}, vol. 34, no. 4, pp. 487--511, 2008.

\bibitem{williams2017hybrid}
Jason~D Williams, Kavosh Asadi, and Geoffrey Zweig,
\newblock ``Hybrid code networks: practical and efficient end-to-end dialog
  control with supervised and reinforcement learning,''
\newblock in {\em ACL}, 2017.

\bibitem{williams2016end}
Jason~D Williams and Geoffrey Zweig,
\newblock ``End-to-end lstm-based dialog control optimized with supervised and
  reinforcement learning,''
\newblock {\em arXiv preprint arXiv:1606.01269}, 2016.

\bibitem{georgila2005learning}
Kallirroi Georgila, James Henderson, and Oliver Lemon,
\newblock ``Learning user simulations for information state update dialogue
  systems.,''
\newblock in {\em Interspeech}, 2005, pp. 893--896.

\bibitem{chandramohan2014co}
Senthilkumar Chandramohan, Matthieu Geist, Fabrice Lefevre, and Olivier
  Pietquin,
\newblock ``Co-adaptation in spoken dialogue systems,''
\newblock in {\em Natural Interaction with Robots, Knowbots and Smartphones},
  pp. 343--353. Springer, 2014.

\bibitem{georgila2014single}
Kallirroi Georgila, Claire Nelson, and David~R Traum,
\newblock ``Single-agent vs. multi-agent techniques for concurrent
  reinforcement learning of negotiation dialogue policies.,''
\newblock in {\em ACL}, 2014, pp. 500--510.

\bibitem{barlier2015human}
Merwan Barlier, Julien Perolat, Romain Laroche, and Olivier Pietquin,
\newblock ``Human-machine dialogue as a stochastic game,''
\newblock in {\em 16th Annual SIGdial Meeting on Discourse and Dialogue
  (SIGDIAL 2015)}, 2015.

\bibitem{hochreiter1997long}
Sepp Hochreiter and J{\"u}rgen Schmidhuber,
\newblock ``Long short-term memory,''
\newblock {\em Neural computation}, vol. 9, no. 8, pp. 1735--1780, 1997.

\bibitem{lee2013structured}
Sungjin Lee,
\newblock ``Structured discriminative model for dialog state tracking,''
\newblock in {\em Proceedings of the SIGDIAL 2013 Conference}, 2013, pp.
  442--451.

\bibitem{henderson2015machine}
Matthew Henderson,
\newblock ``Machine learning for dialog state tracking: A review,''
\newblock in {\em Proc. of The First International Workshop on Machine Learning
  in Spoken Language Processing}, 2015.

\bibitem{williams2016dialog}
Jason Williams, Antoine Raux, and Matthew Henderson,
\newblock ``The dialog state tracking challenge series: A review,''
\newblock {\em Dialogue \& Discourse}, vol. 7, no. 3, pp. 4--33, 2016.

\bibitem{williams1992simple}
Ronald~J Williams,
\newblock ``Simple statistical gradient-following algorithms for connectionist
  reinforcement learning,''
\newblock {\em Machine learning}, vol. 8, no. 3-4, pp. 229--256, 1992.

\bibitem{henderson2014second}
Matthew Henderson, Blaise Thomson, and Jason Williams,
\newblock ``The second dialog state tracking challenge,''
\newblock in {\em SIGDIAL}, 2014.

\bibitem{kingma2014adam}
Diederik Kingma and Jimmy Ba,
\newblock ``Adam: A method for stochastic optimization,''
\newblock in {\em ICLR}, 2014.

\bibitem{srivastava2014dropout}
Nitish Srivastava, Geoffrey~E Hinton, Alex Krizhevsky, Ilya Sutskever, and
  Ruslan Salakhutdinov,
\newblock ``Dropout: a simple way to prevent neural networks from
  overfitting.,''
\newblock {\em Journal of Machine Learning Research}, vol. 15, no. 1, pp.
  1929--1958, 2014.

\bibitem{wen2016conditional}
Tsung-Hsien Wen, Milica Gasic, Nikola Mrksic, Lina~M Rojas-Barahona, Pei-Hao
  Su, Stefan Ultes, David Vandyke, and Steve Young,
\newblock ``Conditional generation and snapshot learning in neural dialogue
  systems,''
\newblock in {\em EMNLP}, 2016.

\end{thebibliography}

\end{document}